\documentclass{article}

\usepackage{cite}
\usepackage{spconf}
\usepackage{amsmath,amssymb,amsfonts}
\usepackage{algorithmic}
\usepackage{graphicx}
\usepackage{textcomp}
\usepackage{xcolor}
\usepackage{svg}
\usepackage[colorlinks=true, allcolors=blue]{hyperref}
\usepackage[nameinlink, capitalise, noabbrev]{cleveref}
\usepackage{supertabular,booktabs}
\usepackage{subcaption}
\usepackage{soul}

\title{Bridging Privacy and Utility: Synthesizing anonymized EEG with constraining utility functions
}
\name{\begin{tabular}{c}Kay Fuhrmeister$^{\star}$ \quad Arne Pelzer$^{\dagger}$ \quad Fabian Radke$^{\dagger}$ \quad Julia Lechinger$^{\ddagger}$ \quad Mahzad Gharleghi$^{\star}$ \\ \textit{Thomas Köllmer$^{\star}$} \quad \textit{Insa Wolf$^{\dagger}$}\end{tabular}}

\address{Fraunhofer Institute for Digital Media Technology IDMT, $^{\star}$Ilmenau / $^{\dagger}$Oldenburg \\
      $^{\ddagger}$University Hospital Schleswig-Holstein, Kiel}

\begin{document}

\maketitle

{\def\thefootnote{}\footnotetext{© 2025 IEEE. Personal use of this material is permitted. Permission from IEEE must be obtained for all other uses, in any current or future media, including reprinting/republishing this material for advertising or promotional purposes, creating new collective works, for resale or redistribution to servers or lists, or reuse of any copyrighted component of this work in other works.}

\begin{abstract}

Electroencephalography (EEG) is widely used for recording brain activity and has seen numerous applications in machine learning, such as detecting sleep stages and neurological disorders. Several studies have successfully shown the potential of EEG data for re-identification and leakage of other personal information. Therefore, the increasing availability of EEG consumer devices raises concerns about user privacy, motivating us to investigate how to safeguard this sensitive data while retaining its utility for EEG applications. To address this challenge, we propose a transformer-based autoencoder to create EEG data that does not allow for subject re-identification while still retaining its utility for specific machine learning tasks. 
We apply our approach to automatic sleep staging by evaluating the re-identification and utility potential of EEG data before and after anonymization. The results show that the re-identifiability of the EEG signal can be substantially reduced while preserving its utility for machine learning.

\end{abstract}


\section{Introduction}
\label{sec:intro}
Electroencephalography (EEG) is a pivotal tool in medicine. It measures the brain’s electrical activity using scalp electrodes. This non-invasive technique is applied in neuroscience, offering valuable insights into cognitive processes, sleep and neurological disorders \cite{EEG_analysis}. 
With artificial intelligence (AI) increasingly utilized in medical research \cite{AI_in_health}, EEG attracts considerable interest in neuroscience due to its rich information content and high temporal resolution \cite{EEG_util}.
However, the sensitive nature of EEG data raises significant concerns regarding the privacy of data subjects and the ethical implications of its use. Regulatory frameworks, such as the Health Insurance Portability and Accountability Act (HIPAA) and the General Data Protection Regulation (GDPR), impose stringent guidelines governing the handling of health-related data. These regulations often restrict data sharing, which stands in conflict with the need for data availability in machine learning (ML) applications. Balancing the need for data with privacy concerns poses a significant obstacle to the development and deployment of innovative AI tools in medicine \cite{big_data_health}.
Sleep medicine, in particular, is a field where the use of AI is promising due to the large amounts of data that need to be gathered and analyzed \cite{sleep_med_ai}, \cite{sleep_med_data}. An important application in sleep medicine is sleep staging, where different stages --wake, N1, N2, N3, REM-- are classified from brain activity, eye movements, and muscle tone \cite{american2007aasm}. This is a time-consuming process where a trained expert manually annotates the sleep stages based on specific patterns in the recorded signals. As such, there is a demand to automate this process using ML applications. The conflict between safeguarding the personal information of data subjects and data availability in such settings underscores the necessity for effective anonymization procedures that can protect the privacy of subjects while preserving the utility of EEG data. This privacy concern is amplified by the increasing availability of personal health devices that are able to record EEG data, like the Muse headband \cite{muse_paper}, \cite{muse_experiment} or research prototypes like the trEEGrid \cite{treegrid}, which are developed for at-home monitoring purposes such as sleep monitoring and could result in widespread EEG data collection. Growing data collection raises privacy risks: EEG recordings can reveal sensitive psychological or neurological information \cite{eeg_anxiety, eeg_psych, eeg_depression, eeg_cognitive, eeg_parkinson}. This is especially concerning in non-clinical settings, where consumer data may receive less protection than in clinical environments. Consequently, developing methods that protect user privacy while retaining data utility is essential.

The first step is to prevent re-identification at the user level. To achieve this, we use a transformer-based autoencoder to create EEG signals that prevent subject re-identification, while still allowing ML-based sleep staging. The main contributions of this paper are:
\begin{itemize}
    \item Implementation of a transformer-based autoencoder to create anonymized EEG signals while retaining utility for ML-based sleep staging.
    \item Application of this approach to automatic sleep staging.
    \item Analysis of the resulting privacy-utility trade-off.
    \item Qualitative analysis of the resulting EEG signals.
\end{itemize}

\section{Related Work}\label{sec:Related_Work}

Transformer-based architectures have already been used for re-identification purposes. In \cite{transformer_reid}, a transformer-based model achieved over $95\%$ accuracy across 109 subjects and multiple test setups with up to $99.8\%$ accuracy in some settings.
Further, EEG-based re-identification has been proposed as a form of user authentication for brain-computer interfaces (BCIs). In \cite{SEHA2022194}, the authors created augmented embeddings from users' EEG signals, yielding robust re-identification for a group of 50 users. In another test setting with 12 users, the re-identification remained robust between EEG signals recorded almost one year apart when using 10 seconds of signal. The use of EEG-based BCI user authentication has received a considerable amount of attention. In \cite{bci_review}, the authors found 68 different academic works based on 108 distinct experiments between 1998 and 2022.
In comparison, privacy protection of EEG data remains comparatively understudied.

One method for user identity anonymization for EEG-based BCIs was proposed in \cite{bci}. The authors created privacy-protecting EEG signals by injecting perturbation into the signal. The perturbation is created through a ML-based approach which generates noise that correlates with each BCI user and then adds all user-specific noise signals to the original EEG data, thus obscuring user identity.
Further, an autoencoder approach to anonymize EEG signals has been proposed in \cite{autoencoder}. The authors used a CNN autoencoder to selectively anonymize EEG signals while preserving the utility for stimuli classification tasks. The approach showed promising results for anonymization and utility preservation.
However, these works heavily rely on CNN architectures and do not further analyze leftover re-identification potential remaining in the anonymized data.

\section{Experiments}\label{sec:Experiments}
For the anonymization of EEG data, we implement a transformer-based autoencoder inspired by \cite{autoencoder} that takes EEG data as input and outputs a signal optimized to retain its performance for a given utility case while removing identifying characteristics of the original signal. Our experiments focus on the utility case of sleep staging.

\subsection{Setup}\label{sec:Setup}
The setup to train our autoencoder includes three components -- the utility model, the re-identification model and the autoencoder for the anonymization of the EEG signal. 
We compare results for two separate utility model architectures. We use a self-trained DeepSleepNet \cite{deepsleep} and a pretrained RobustSleepNet \cite{robustsleepnet} obtained from the authors of \cite{robust_sleep_oldenburg}. RobustSleepNet was pretrained on EEG and electrooculographic (EOG) data but only receives EEG during inference in our experiments, which lowers performance, especially for the REM stage. Since we are mainly interested in relative differences between non-anonymized and anonymized EEG, this is deemed acceptable.
The utility and the re-identification models are both trained on data from Sleep-EDF \cite{sleep-edf}, except for RobustSleepNet which was pre-trained on MASS \cite{mass}. 
The re-identification model is a transformer-based classifier that takes EEG inputs in the same shape as the utility model's inputs and predicts a class label corresponding to a subject in the dataset.
The autoencoder takes an EEG signal as input and creates an anonymized output signal used as input for the utility and the re-identification models. The autoencoder is based on a transformer architecture and consists of four transformer encoder layers followed by four transformer decoder layers, each with eight attention heads. An overview of the setup is shown in \Cref{fig:setup}. 

\begin{figure}[h!]
\centering
\includegraphics[scale=0.35]{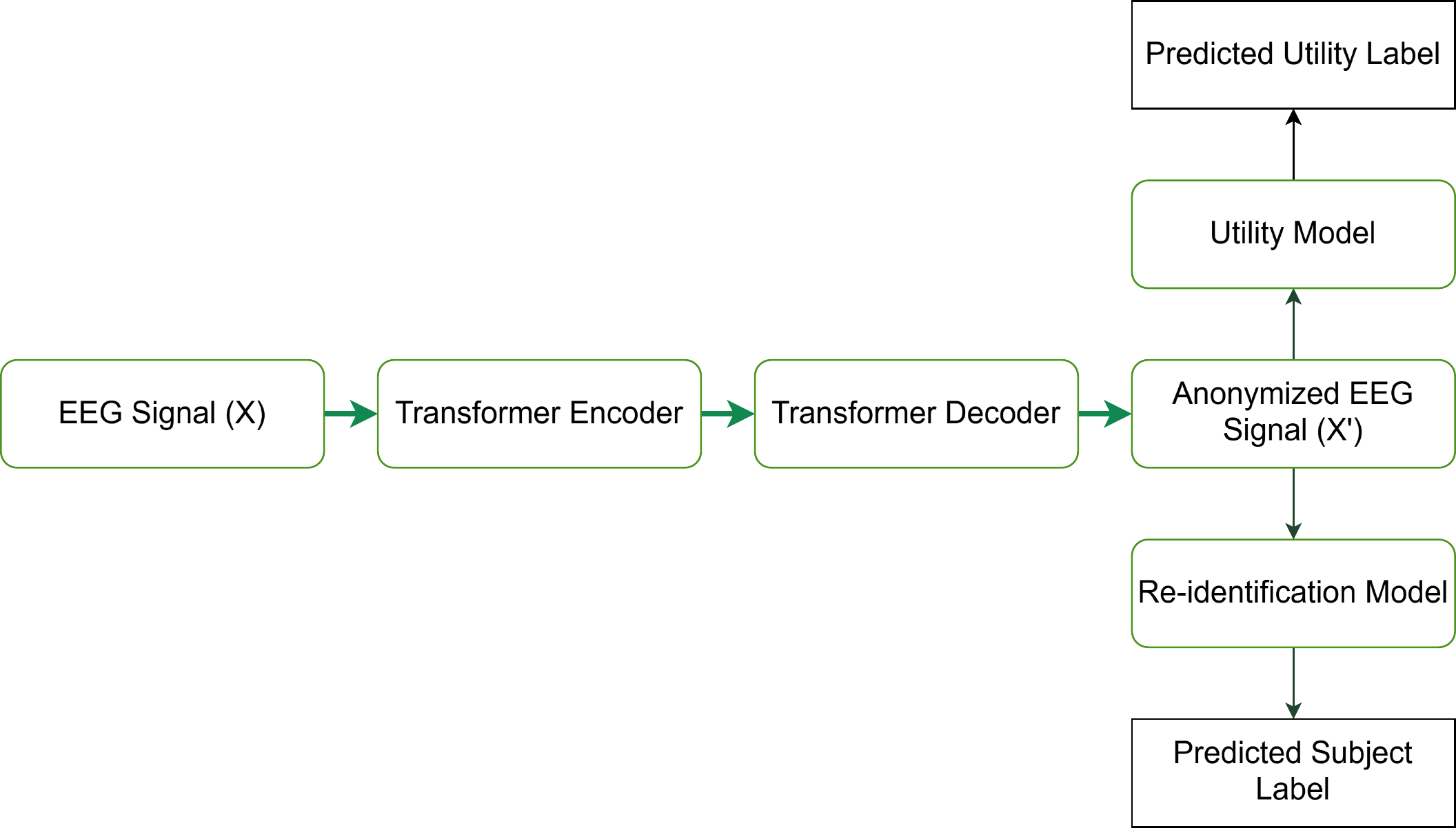}
\caption{Setup for training the autoencoder.}
\label{fig:setup}
\end{figure}
The autoencoder is trained with a combined loss function $L$ from the utility model's loss $L_{util}$ and re-identification model's loss $L_{Id}$. We set
\begin{align}
L = \omega_{util} \cdot L_{util} - \omega_{Id} \cdot  L_{Id} + \omega_{dist} \cdot L_{dist},\label{loss}
\end{align}
where $L_{dist}$ is a suitable distance metric. This parameter is added to incentivize an output signal that preserves the basic shape of the EEG inputs. For this, we choose the mean squared error (MSE) between the input and output. The weights $\omega_*$ give different emphasis on the respective loss components. During the training process, only the parameters of the autoencoder are optimized. The parameters of the utility model and the re-identification model remain fixed. This is done to ensure that the autoencoder learns to produce anonymized outputs, instead of the re-identification model adjusting its weights to perform poorly. For sleep staging with DeepSleepNet, the autoencoder was trained for 30 epochs with a batch size of 32, a learning rate of $4\cdot 10^{-6}$ and weights $\omega_{util} = 2000, \, \omega_{Id} = 25,\, \omega_{dist}=1$. For RobustSleepNet, the autoencoder was trained with a learning rate of $8\cdot 10^{-7}$, weights $\omega_{util} = 1300, \, \omega_{Id} = 10,\, \omega_{dist}=1$ and a batch size of $64$.

The training set for the autoencoder is created using a stratified split, with $80\%$ of the data being used for training and the remaining $20\%$ being used as a test set. 
After training the autoencoder, the utility and re-identification models are evaluated on an anonymized version of the test set. These results are compared to the base performance on original data. Additionally, we used a trained autoencoder to create an anonymized version of our dataset. Then a fresh DeepSleepNet and re-identification model are trained and evaluated on said data. This is to check the viability of the anonymized data for downstream tasks and to estimate how much potentially identifiable information remains in the anonymized signal.

\subsection{Data}\label{sec:Data}
The experiments were carried out on the sleep cassette study (SC) of the expanded Sleep-EDF dataset \cite{sleep-edf}, which we will refer to as SleepEDFx-SC. 
The data includes a total of 78 healthy subjects aged between 25 and 101. The recordings were taken at the subjects' homes for two subsequent day-night periods and contain two EEG channels. EEG signals were sampled at 100 Hz and annotated by trained technicians following the Rechtschaffen and Kales manual \cite{sleep_scoring}. We adapt these annotations to the AASM guidelines \cite{american2007aasm} by merging stages 3 and 4 into N3. For our experiments, we use a subset of 35 younger subjects between the ages of 25 to 60. We extracted the sleep portion of the SleepEDFx-SC recordings by identifying the onset of the first segment labeled as a sleep stage (N1) and the end of the last segment annotated as a sleep stage. To ensure the presence of wakefulness, 30 minutes were added before the onset and after the end of the sleep period. The extracted recordings were then manually checked to ensure correctness.

\section{Evaluation}\label{sec:Evaluation}
The results show a substantial reduction in the re-identification accuracy on anonymized data. The baseline accuracy for re-identification for SleepEDFx-SC is $0.86$. After anonymization, the pre-trained model is re-evaluated on anonymized data and only reaches an accuracy of $0.030$, which is close to the guessing probability of $0.029$. This is in line with previous results in \cite{autoencoder}, however, further evaluations were not carried out by the authors. Additionally, we re-evaluate the re-identification with a fresh model trained from scratch on an anonymized version of the SleepEDFx-SC data. For the anonymization for DeepSleepNet, re-identification accuracy still reaches $0.65$, indicating a substantial decrease in identifiability while also signaling the remaining presence of subject-specific information in the anonymized data. 
For RobustSleepNet, the pre-trained re-identification model only reaches an accuracy of $0.01$ and a fresh re-identification model trained on anonymized data reaches $0.62$. The subject-wise re-identification results for both utility models are shown in \Cref{fig:reid_sleep}.
\begin{figure}[h]
\centering
\centerline{\includegraphics[scale=0.29]{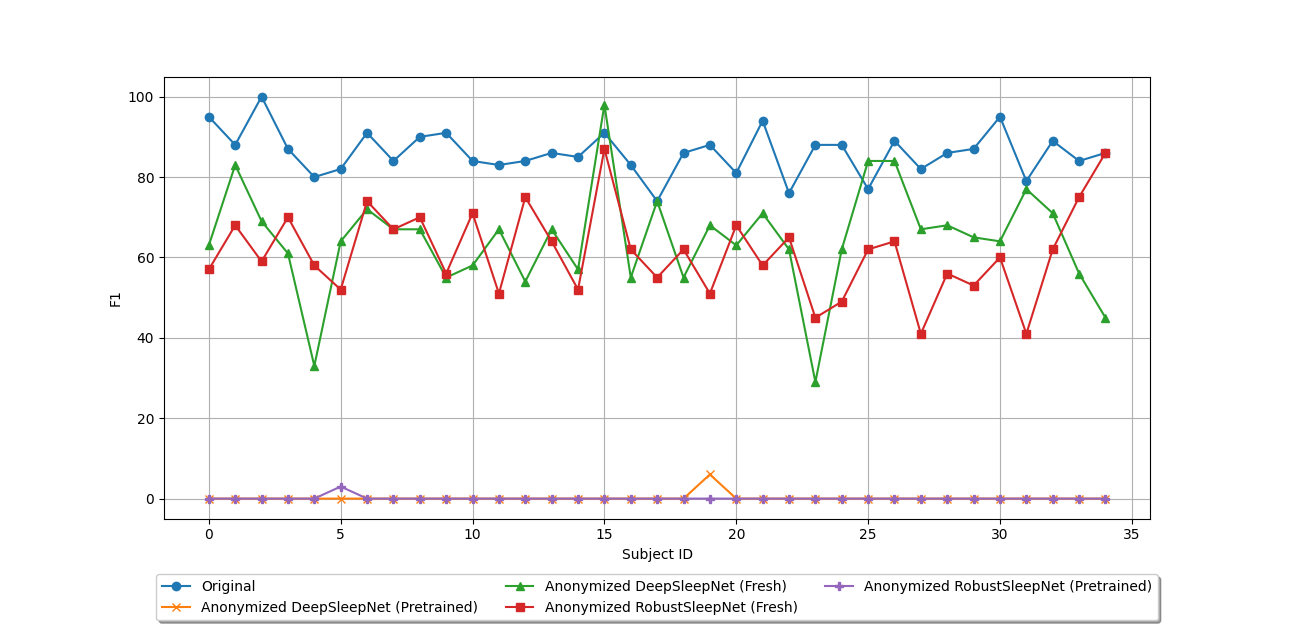}}
\caption{Comparison of re-identification accuracies using subject-wise F1-score for each subject in the dataset.}
\label{fig:reid_sleep}
\end{figure}
The figure shows that there is a lot of variance among the subject-wise F1-scores. In particular, some outliers can be identified much better than others. Concerning the utility of the data, the overall accuracy on the original data reaches $0.78$ for DeepSleepNet and $0.68$ RobustSleepNet. After anonymization, there is a slight decrease in performance for both models. The accuracy drops to $0.76$ for DeepSleepNet and $0.65$ for RobustSleepNet. These evaluations are carried out again for DeepSleepNet by training and evaluating a fresh model on an anonymized version of the SleepEDFx-SC data, with an accuracy of $0.77$. The utility results are summarized in \cref{tab:summary}, showing F1-scores per sleep stage and overall accuracy.

\begin{table}[h!]
    \centering
    \caption{Summary of the utility evaluation results.}
    \begin{tabular}{c|c|c|c|c|c|c}
    Evaluation & Wake & N1 & N2 & N3 & REM & Acc \\
    \hline
    DeepSleep (orig) & 0.90 & 0.36 & 0.77 & 0.80 & 0.75 & 0.78 \\
    DeepSleep (anon) & 0.83 & 0.30 & 0.83 & 0.82 & 0.67 & 0.76 \\
    DeepSleep (anon fresh) & 0.91 & 0.32 & 0.82 & 0.81 & 0.67 & 0.77 \\
    RobustSleep (orig) & 0.86 & 0.25 & 0.71 & 0.62 & 0.17 & 0.68 \\
    RobustSleep (anon) & 0.82 & 0.18 & 0.74 & 0.60 & 0.31 & 0.65 
    \end{tabular}
    \label{tab:summary}
\end{table}

\section{Discussion}\label{sec:Discussion}

The results show the possibility of removing personally identifying data from EEG signals. Both experiments show that a model trained on original data completely fails to recognize anonymized data. However, fresh models trained on anonymized data show the remaining presence of identifiable information.
For DeepSleepNet, the corresponding drop in the re-identification accuracy is $21\%$ compared to original data, while resulting only in a slight drop in utility accuracy of $1\%$. 
In comparison, the performance of RobustSleepNet drops by $3\%$, from $68\%$ to $65\%$. On the other hand, the anonymization is slightly stronger, with re-identification accuracy dropping by $24\%$. 
In addition to removing personal data, the autoencoder can preserve general patterns of the EEG data in its outputs.
A sample segment and its corresponding anonymized versions are shown in \Cref{fig:eeg_samples}. It shows the original segment along with two anonymized counterparts, one anonymized for DeepSleepNet, the other for RobustSleepNet.
\begin{figure}[ht]
\begin{subfigure}{.5\textwidth}
    \centering
    \includegraphics[scale=0.15]{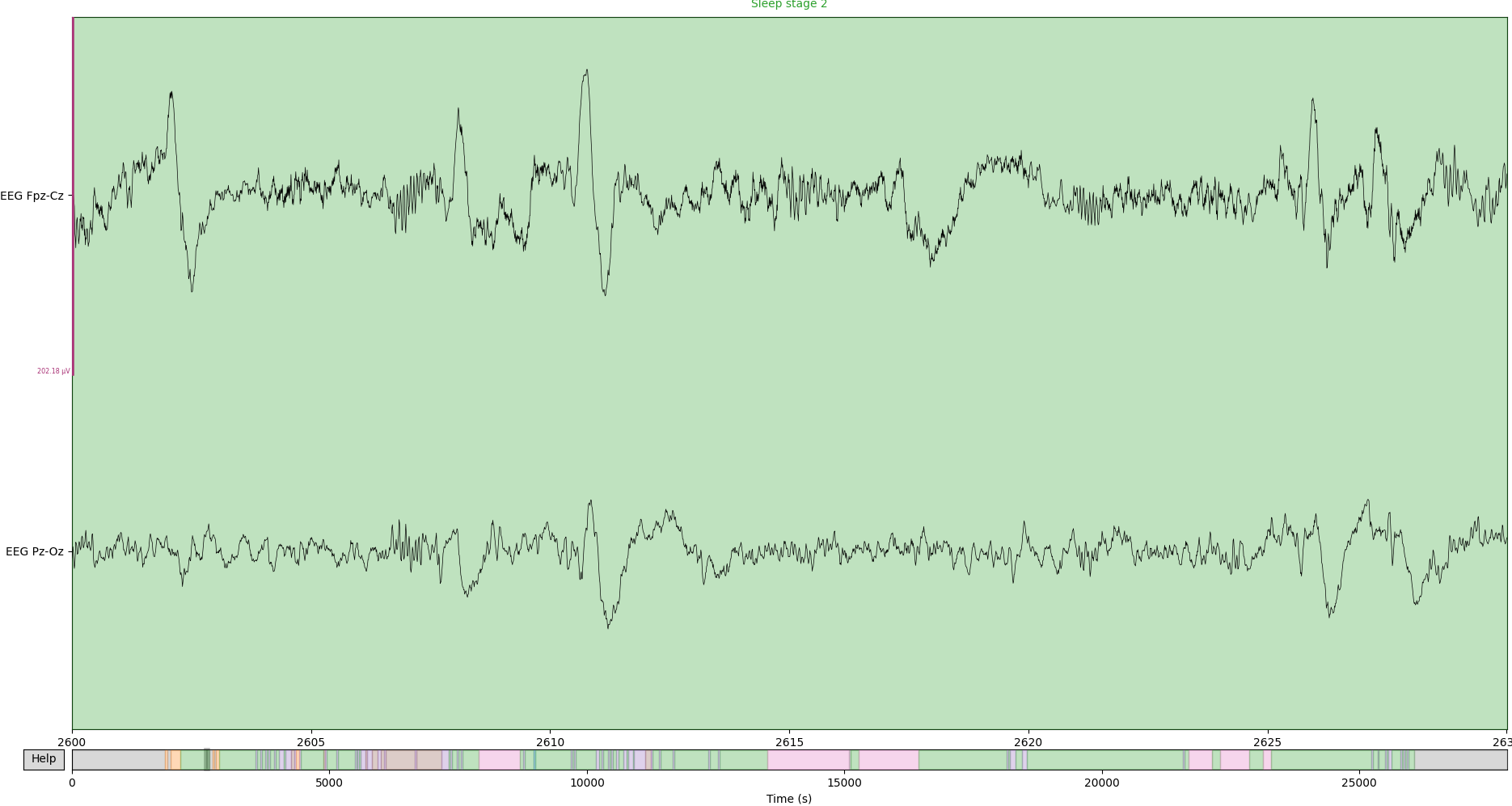}
    \caption{Original EEG segment.}
    \label{raw_eeg}
\end{subfigure}
\begin{subfigure}{.5\textwidth}
    \centering
    \includegraphics[scale=0.15]{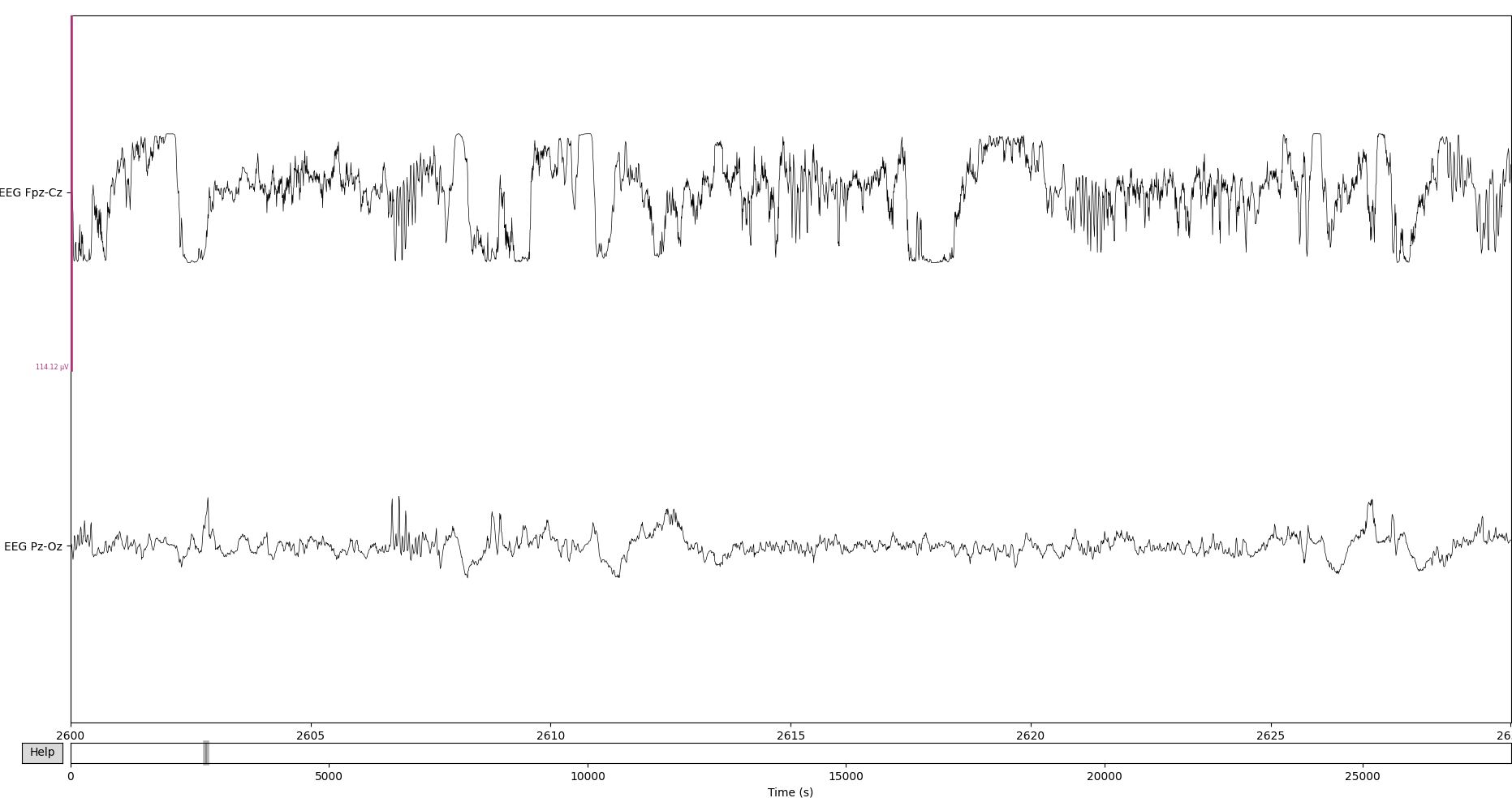}
    \caption{EEG segment anonymized for DeepSleepNet.}
    \label{anon_eeg_sleep}
\end{subfigure}
\begin{subfigure}{.5\textwidth}
    \centering
    \includegraphics[scale=0.15]{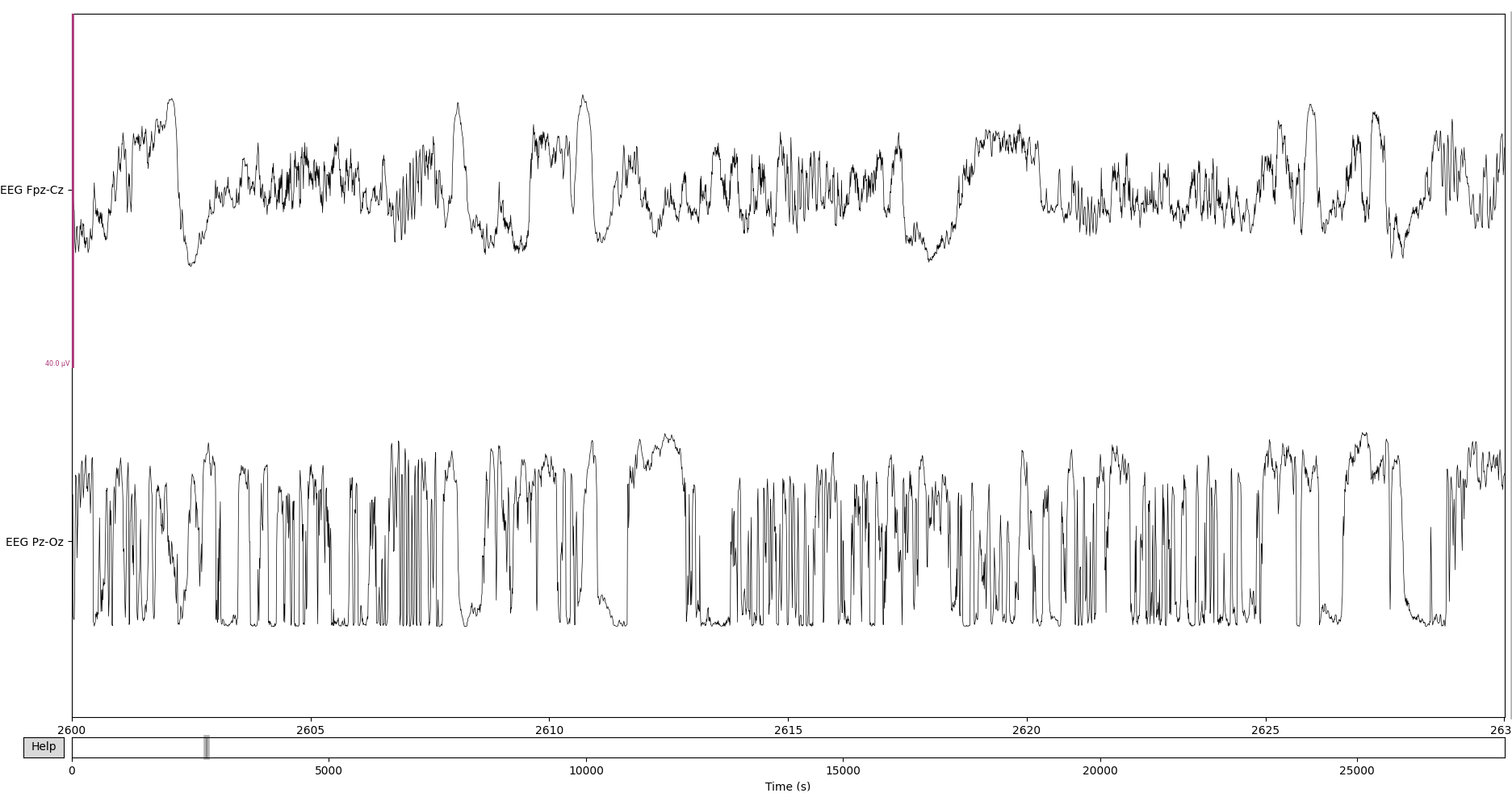}
    \caption{EEG segment anonymized for RobustSleepNet.}
    \label{anon_eeg_sleep_robust}
\end{subfigure}
\caption{Original EEG segment and anonymized version.}
\label{fig:eeg_samples}
\end{figure}
In the segment anonymized for DeepSleepNet, rough patterns from the original data are still visible. For RobustSleepNet, the bottom channel seems more distorted with high frequency patterns, while the top one shows more similarities with the original signal.
DeepSleepNet preserves more geometric features in the data, likely due to its use of convolutional layers. In comparison, RobustSleepNet uses attention in the time domain of the EEG signal. This could enable the signal to retain utility through different features in the signal, leading to different output shapes.
Moreover, the power spectral densities retain similarities to the original, although the two graphs belonging to the two channels are more separated. For RobustSleepNet, the curve is within a lower value range. An overview is given in \Cref{fig:psd_compare}.
\begin{figure}[h!]
\begin{subfigure}{.5\textwidth}
    \centering
    \includegraphics[trim={0cm 0cm 0cm 0.6cm},clip, scale=0.3]{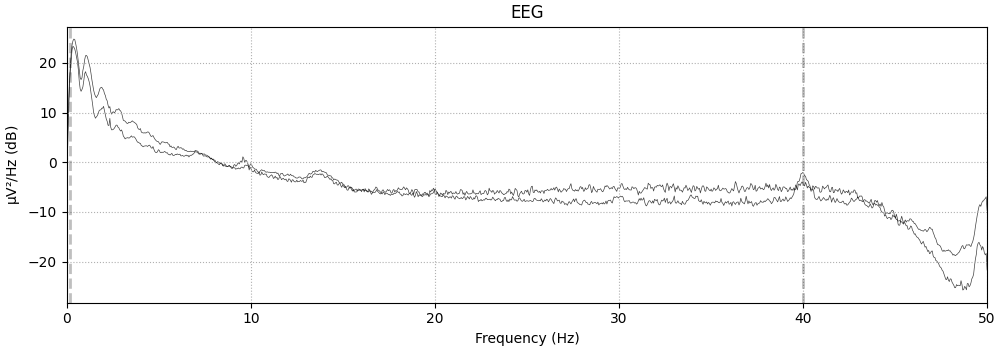}
    \caption{PSD of original EEG.}
    \label{psd_raw_eeg}
\end{subfigure}
\begin{subfigure}{.5\textwidth}
    \centering
    \includegraphics[trim={0cm 0cm 0cm 0.6cm},clip,scale=0.3]{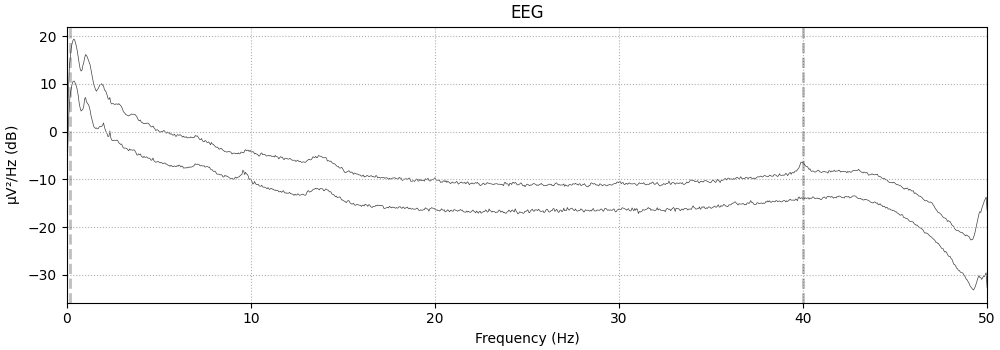}
    \caption{PSD for EEG anonymized for DeepSleepNet.}
    \label{anon_psd_sleep}
\end{subfigure}
\begin{subfigure}{.5\textwidth}
    \centering
    \includegraphics[trim={0cm 0cm 0cm 0.6cm},clip,scale=0.3]{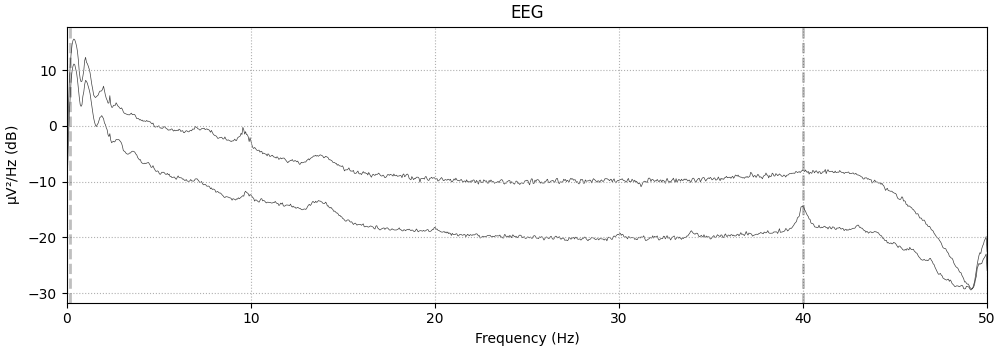}
    \caption{PSD for EEG anonymized for RobustSleepNet.}
    \label{anon_psd_sleep_robust}
\end{subfigure}
\caption{Comparison of PSDs for original and anonymized EEG recordings. Filtered between 0.2 and 40 Hz.}
\label{fig:psd_compare}
\end{figure}
Additionally, there are outliers in all re-identification experiments. The results show that there is a risk of certain subjects being considerably more identifiable than others. The six most identifiable subjects for DeepSleepNet are three females, aged 30, 36 and 51 as well as three males aged 28, 54 and 60. 
The three most notable outliers for RobustSleepNet are two female aged 54 and 56, and one male, aged 54. However, the subjects belonging to the IDs 30 and 32 are two females aged 54 and 56 and show no signs of high identifiability.
While older subjects make up the majority of the outliers, there are no clear physiological similarities between all of them. Further, environmental influences or device specific noise could be responsible for the increased identifiability of the subjects. The subjects' EEG data was recorded in an at-home setting instead of a controlled environment, hence, there are a lot of potential causes for this. This includes device-specific signal noise, device interference 
or unique stimuli in the home environment. 
Notably, outliers vary depending on the utility model used in the training process. Therefore, their presence could indicate some model-specific overfitting, or a strong correlation between some subject-specific features and important features for the utility model's predictions. A further investigation of these outliers should be conducted in future work. For the time being, this approach could be used to identify individuals who are at high risk for re-identification and preemptively remove them from data releases. 
We are also planning to further investigate the utility of the produced anonymized data with clinicians to assess possible future use cases for our approach.

\section{Conclusion}\label{sec:conclusion}
In this paper, we have successfully demonstrated the potential of using a transformer-based autoencoder architecture to remove personal identifying information from EEG data while still retaining its utility for the ML task of sleep staging. We have further analyzed how results vary depending on the used utility model. Additionally, we analyzed the continued persistence of personal information in the anonymized data afterwards.

\section*{\textbf{Acknowledgment}}
This research was conducted as part of the NEMO project funded by the Federal Ministry of Research, Technology and Space (BMFTR). We would like to extend our sincere gratitude to Robert Göder and Clint Hansen from the University Hostipal Schleswig-Holstein for their valuable insights and support throughout this study.

\bibliographystyle{ieeetr}

{\small
\bibliography{bibliography}}

\begin{thebibliography}{10}

\bibitem{EEG_analysis}
H.~Zhang, Q.-Q. Zhou, H.~Chen, X.-Q. Hu, W.-G. Li, Y.~Bai, J.-X. Han, Y.~Wang, Z.-H. Liang, D.~Chen, F.-Y. Cong, J.-Q. Yan, and X.-L. Li, ``The applied principles of eeg analysis methods in neuroscience and clinical neurology,'' {\em Military Medical Research}, vol.~10, 2023.

\bibitem{AI_in_health}
J.~Bajwa, U.~Munir, A.~Nori, and B.~Williams, ``Artificial intelligence in healthcare: transforming the practice of medicine,'' {\em Future Healthcare Journal}, vol.~8, no.~2, pp.~e188--e194, 2021.

\bibitem{EEG_util}
C.~M. Michel and M.~M. Murray, ``Towards the utilization of eeg as a brain imaging tool,'' {\em NeuroImage}, vol.~61, no.~2, pp.~371--385, 2012.
\newblock Neuroimaging: Then, Now and the Future.

\bibitem{big_data_health}
R.~Pastorino, C.~De~Vito, G.~Migliara, K.~Glocker, I.~Binenbaum, W.~Ricciardi, and S.~Boccia, ``Benefits and challenges of big data in healthcare: an overview of the european initiatives,'' {\em European Journal of Public Health}, vol.~29, 11 2019.

\bibitem{sleep_med_ai}
N.~F. Watson and C.~R. Fernandez, ``Artificial intelligence and sleep: Advancing sleep medicine,'' {\em Sleep Medicine Reviews}, vol.~59, p.~101512, 2021.

\bibitem{sleep_med_data}
I.~Perez-Pozuelo, B.~Zhai, J.~Palotti, R.~Mall, M.~Aupetit, J.~M. Garcia-Gomez, S.~Taheri, Y.~Guan, and L.~Fernandez-Luque, ``The future of sleep health: a data-driven revolution in sleep science and medicine,'' {\em npj Digital Medicine}, vol.~3, 2020.

\bibitem{american2007aasm}
A.~A. of~Sleep~Medicine {\em et~al.}, ``The aasm manual for the scoring of sleep and associated events: rules, terminology and technical specifications,'' {\em Westchester, IL: American Academy of Sleep Medicine}, vol.~23, 2007.

\bibitem{muse_paper}
N.~Kovacevic, P.~Ritter, W.~Tays, S.~Moreno, and A.~R. McIntosh, ``‘my virtual dream’: Collective neurofeedback in an immersive art environment,'' {\em PLOS ONE}, vol.~10, pp.~1--18, 07 2015.

\bibitem{muse_experiment}
O.~E. Krigolson, C.~C. Williams, A.~Norton, C.~D. Hassall, and F.~L. Colino, ``Choosing muse: Validation of a low-cost, portable eeg system for erp research,'' {\em Frontiers in Neuroscience}, vol.~Volume 11 - 2017, 2017.

\bibitem{treegrid}
C.~F. da~Silva~Souto, W.~Pätzold, M.~Paul, S.~Debener, and K.~I. Wolf, ``Pre-gelled electrode grid for self-applied eeg sleep monitoring at home,'' {\em Frontiers in Neuroscience}, vol.~Volume 16 - 2022, 2022.

\bibitem{eeg_anxiety}
A.~Al-Ezzi, N.~Yahya, N.~Kamel, I.~Faye, K.~Alsaih, and E.~Gunaseli, ``Severity assessment of social anxiety disorder using deep learning models on brain effective connectivity,'' {\em IEEE Access}, vol.~9, pp.~86899--86913, 2021.

\bibitem{eeg_psych}
S.~M. Park, B.~Jeong, D.~Y. Oh, C.-H. Choi, H.~Y. Jung, J.-Y. Lee, D.~Lee, and J.-S. Choi, ``Identification of major psychiatric disorders from resting-state electroencephalography using a machine learning approach,'' {\em Frontiers in Psychiatry}, vol.~Volume 12 - 2021, 2021.

\bibitem{eeg_depression}
A.~Rafiei and Y.-K. Wang, ``Automated major depressive disorder classification using deep convolutional neural networks and choquet fuzzy integral fusion,'' in {\em 2022 IEEE Symposium Series on Computational Intelligence (SSCI)}, pp.~186--192, 2022.

\bibitem{eeg_cognitive}
M.~Aljalal, M.~Molinas, S.~A. Aldosari, K.~AlSharabi, A.~M. Abdurraqeeb, and F.~A. Alturki, ``Mild cognitive impairment detection with optimally selected eeg channels based on variational mode decomposition and supervised machine learning,'' {\em Biomedical Signal Processing and Control}, vol.~87, p.~105462, 2024.

\bibitem{eeg_parkinson}
M.~Nour, U.~Senturk, and K.~Polat, ``Diagnosis and classification of parkinson's disease using ensemble learning and 1d-pdcovnn,'' {\em Computers in Biology and Medicine}, vol.~161, p.~107031, 2023.

\bibitem{transformer_reid}
Y.~Du, Y.~Xu, X.~Wang, L.~Liu, and P.~Ma, ``Eeg temporal–spatial transformer for person identification,'' {\em Scientific Reports}, vol.~12, 2022.

\bibitem{SEHA2022194}
S.~N.~A. Seha and D.~Hatzinakos, ``A new training approach for deep learning in eeg biometrics using triplet loss and emg-driven additive data augmentation,'' {\em Neurocomputing}, vol.~488, pp.~194--211, 2022.

\bibitem{bci_review}
C.~A. Fidas and D.~Lyras, ``A review of eeg-based user authentication: Trends and future research directions,'' {\em IEEE Access}, vol.~11, pp.~22917--22934, 2023.

\bibitem{bci}
L.~Meng, X.~Jiang, J.~Huang, W.~Li, H.~Luo, and D.~Wu, ``User identity protection in eeg-based brain–computer interfaces,'' {\em IEEE Transactions on Neural Systems and Rehabilitation Engineering}, vol.~31, pp.~3576--3586, 2023.

\bibitem{autoencoder}
G.~e.~a. Singh, ``Selective eeg signal anonymization using multi-objective autoencoders,'' in {\em 2023 20th Annual International Conference on Privacy, Security and Trust (PST)}, pp.~1--7, 2023.

\bibitem{deepsleep}
A.~Supratak, H.~Dong, C.~Wu, and Y.~Guo, ``Deepsleepnet: A model for automatic sleep stage scoring based on raw single-channel eeg,'' {\em IEEE Transactions on Neural Systems and Rehabilitation Engineering}, vol.~25, no.~11, pp.~1998--2008, 2017.

\bibitem{robustsleepnet}
A.~Guillot and V.~Thorey, ``Robustsleepnet: Transfer learning for automated sleep staging at scale,'' {\em IEEE Transactions on Neural Systems and Rehabilitation Engineering}, vol.~29, pp.~1441--1451, 2021.

\bibitem{robust_sleep_oldenburg}
F.~A. Radke, C.~F. da~Silva~Souto, W.~Pätzold, and K.~I. Wolf, ``Transfer learning for automatic sleep staging using a pre-gelled electrode grid,'' {\em Diagnostics}, vol.~14, no.~9, 2024.

\bibitem{sleep-edf}
B.~Kemp, A.~Zwinderman, B.~Tuk, H.~Kamphuisen, and J.~Oberye, ``Analysis of a sleep-dependent neuronal feedback loop: the slow-wave microcontinuity of the eeg,'' {\em IEEE Transactions on Biomedical Engineering}, vol.~47, no.~9, pp.~1185--1194, 2000.

\bibitem{mass}
C.~O'Reilly, N.~Gosselin, J.~Carrier, and T.~Nielsen, ``Montreal archive of sleep studies: an open-access resource for instrument benchmarking and exploratory research,'' {\em Journal of Sleep Research}, vol.~23, no.~6, pp.~628--635, 2014.

\bibitem{sleep_scoring}
A.~Rechtschaffen and A.~Kales, {\em A manual of standardized terminology, techniques and scoring systems for sleep stages of human subjects}.
\newblock Public Health Service, US Government Printing Office, Washington DC, 1968.

\end{thebibliography}

\end{document}